\documentclass[conference]{IEEEtran}
\IEEEoverridecommandlockouts
\usepackage{cite}
\usepackage{amsmath,amssymb,amsfonts}
\usepackage{algorithmic}
\usepackage{graphicx}
\usepackage{subfigure}
\usepackage{textcomp}
\usepackage{xcolor}

\def\BibTeX{{\rm B\kern-.05em{\sc i\kern-.025em b}\kern-.08em
    T\kern-.1667em\lower.7ex\hbox{E}\kern-.125emX}}
\begin{document}

\title{EOLO: Embedded Object Segmentation only Look Once\\
}

\author{\IEEEauthorblockN{Longfei Zeng}
\IEEEauthorblockA{\textit{Lakehead University} \\
Thunder Bay, Canada \\
lzeng3@lakeheadu.ca}
\and
\IEEEauthorblockN{Sabah Mohammed}
\IEEEauthorblockA{\textit{dept. Computer Science, Lakehead University} \\
Thunder Bay, Canada \\
mohammed@lakeheadu.ca}

}

\maketitle

\begin{abstract}
In this paper, we introduce an anchor-free and single-shot instance segmentation method, which is conceptually simple with 3 independent branches, fully convolutional and can be used by easily embedding it into mobile and embedded devices.

Our method, refer as EOLO, reformulates the instance segmentation problem as predicting semantic segmentation and distinguishing overlapping objects problem, through instance center classification and 4D distance regression on each pixel. Moreover, we propose one effective loss function to deal with sampling high-quality center of gravity examples and optimization for 4D distance regression, which can significantly improve the mAP performance. Without any bells and whistles, EOLO achieves 27.7$\%$ in mask mAP under IoU50 and reaches 30 FPS on 1080Ti GPU, with single-model and single-scale training/testing on the challenging COCO2017 dataset.

For the first time, we show the different comprehension of instance segmentation in recent methods, in terms of both up-bottom, down-up, and direct-predict paradigms. Then we illustrate our model and present related experiments and results. We hope that the proposed EOLO framework can serve as a fundamental baseline for a single-shot instance segmentation task in Real-time Industrial Scenarios.
\end{abstract}


\section{Introduction}
Instance segmentation is a more complex task comparing with object detection and semantic segmentation. It requires predicting each instance not only an approximate location but also a pixel-level segmentation. The recent instance segmentation networks tend to be lighter and try to keep the State-of-the-Art performance. Despite the anchor-free and one-stage detectors have promoted the speed of inference, these advanced algorithms are not small enough and inference slow for most industrial application scenarios. It is still a challenge to implement a faster and smaller instance segmentation network on a computationally limited platform. To break through this dilemma, this paper proposes an efficient and succinct instance segmentation network for embedded vision application scenarios.

There are four categorizations of instance segmentation algorithms, two-stage or one-stage paradigms, and top-down or bottom-up paradigms. Mask R-CNN\cite{MaskR-CNN} and its' relative derivatives are following top-down and two-stage paradigms. It first detects objects by bounding boxes and classification then fine-tuning the bounding boxes and segments the instance mask in each bounding box. TWO-stage paradigm improved 
accuracy but the dependant branches and luxurious computation decide the difficulties of real-time, it is impossible to deploy Mask R-CNN on an embedded platform. 

The recent computer vision algorithms use simple pipelines of one-stage. One-stage instance segmentation is affected by one-stage target detection research, such as early anchor-based detection model YOLO\cite{YOLO} and RetinaNet\cite{RetinaNet}, recent anchor-free detection models like FCOS\cite{FCOS} and CenterNet\cite{CenterNet}. The design of anchors limited the generalization of the model,  the model cannot adapt to new data, new scenarios, and new scales. Moreover anchors also increased the computation volume. In the industry should know that these anchor-based models are not practical. Recently, anchor-free detectors can outperform the anchor-base detectors in both accuracy and computation volume. Anchor-free and one-stage have gradually become the industry's better even the best choice.

\begin{figure}[htp]
    \centering
    \includegraphics[width=8.5cm]{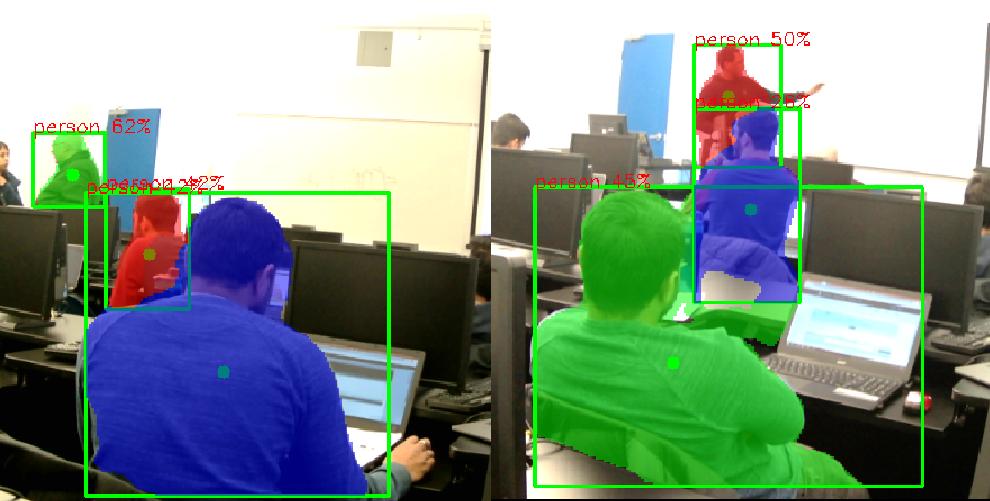}
    \caption{The EOLO results on real data collected from Raspberry Pi4 camera accelerate with Google Coral USB Accelerator, reaches 18 FPS for inference.}
\label{structure}
\end{figure}

In fundamentally, top-down and bottom-up methods both are exploring the relationship between objects and pixels (relationship between semantic and pixels). Recently advanced one-stage and anchor-free instance segmentation such as PolarMask\cite{PolarMask} provides a direct answer, PolarMask describes contours using 36 points which are the intersections of 36 fixed-direction rays from the object center to object boundary, the center point on the feature map is represented by a 2D vector-Center(x, y),  each center has 36 distance values which could describe the points on the contours of the instance. This method converts the instance division problem into instance center point classification problem and dense distance regression problem.  Compared to FCOS, which extends radiates rays from 4D (from the center to left, right, top and bottom) to 36D (from the center to 36 fixed directions). Instance segmentation is transferred as an object detection problem. Because of the shape diversity, PolarMask could not fig out the concave and irregular polygon problems. Increases the counts of rays could relieve this problem, but cannot solve it. Increasing counts of rays will increase the computation volume relative original 36-values regression problem.

SOLO\cite{SOLO} is another one-stage and anchor-free bottom-up method, it proposes an innovative way to represent the relationship between semantic and pixels. The author points out that the instance segmentation is the processing detecting the center of the object (position) and the object's size. SOLO divides a picture into an S × S grid, which represents S * S positions on the picture. Different from TensorMask\cite{TensorMask}, the SOLO project each pixels' position information on the channel dimension of the feature map. The SOLO idea refers to semantic segmentation, classifies the pixels to the affiliated center point. So the information on the geometric structure is retained. A location prediction problem is transformed from a regression problem to a classification problem. The significance of this method is an intuitive and simple classification method, it models a variable number of instances with a fixed number of channels without relying on post-processing methods. But this classification task is a heavy task, both shape of feature map H*W and position map S*S should be appropriately token as a large value to fit small scale instances. This method increases the computation volume to a significant level.

The Instance Segmentation on embedded devices question is converted to "Is there a light cost way to build the relationship of pixels and instances?" Actually, according to the research, we found using 4 values is enough to express the relationship between pixels and its instance. We extended the function of object size to help model understand Object overlapping parts. With the proposed EOLO framework, we are able to implement the instance segmentation in an end-to-end fashion with less post-processing.

\section{Related Work}
We review some object detection and instance segmentation works that are relative to our work.

\subsection{Anchor-free Object Detection}\label{AA}
The anchor-base serious methods like R-CNN\cite{RCNN}, Fast R-CNN\cite{FastRCNN}, Faster R-CNN\cite{FasterRCNN}, YOLOv2, v3\cite{yolov2}\cite{yolov3}, SSD\cite{ssd} rely on the advance statistic for anchors shape, this processing limits the generalizations of models. Additionally, R-CNN derivatives depend on region proposal methods to extract interesting regions. Those methods are time wasteful. Recently, anchor-free networks have achieved dramatic success in computer vision tasks like object detection\cite{DenseBox},\cite{RetinaNet},\cite{FCOS},\cite{RepPoints},\cite{CenterNet},\cite{Keypoint},\cite{FoveaBox}, instance segmentation\cite{yolact},\cite{TensorMask},\cite{Bottom-up}. As a basis of instance segmentation, the success of anchor-free object detection promotes the instance segmentation. YOLOv1 has been an anchor-free detector, it divides feature map as 14*14 grids, and predicts bounding boxes and center of objects at the same time. However, YOLOv2 and YOLOv3 have been anchor-based detector. The CenterNet is one of the successful anchor-free object detectors. It uses keypoint estimation minding to predict center points and directly regress all objects size which is represented by a 2D vector (W, H) to build a bounding box. It avoids to process a huge set of region candidates and avoid calculating the intersection-over-union (IoU) during training. The CenterNet uses max pooling to extract the peek points on the center heatmap, which saves competition from non-maximum suppression (NMS) post-processing. It is worth mentioning that, CneterNet does not deploy multilevel prediction to solve multilevel scales problem, it works on 1/4 size of input image heatmap, 1/4 resolution could grantee small, middle and large scale object simultaneously detected.
 
 FCOS is an FCN-based pixel-by-pixel target detection algorithm, which implements anchor-free and proposal free solutions, and proposes the idea of Center-Ness. The performance of the recall rate is better than many advanced anchor-based object detection algorithms. This algorithm defines bounding box by a 4D vector, it works on predicted center and left, top, right and bottom distances from the center to the bounding box.  The processing of the regression 4D vector happens on each pixel of the heatmap. The Center-Ness layer makes sure reliable regression results that are close to the center,  mainly contribute to the final bounding box of objects. The 4D vector can not only contribute to the object size, but it can also help to distinguish the overlapping part between two or more objects. In this essay, we trained part of those points except the center point to build up the relationship between pixels to objects. 
 
\begin{figure*}[htp]
    \centering
    \includegraphics[width=18cm]{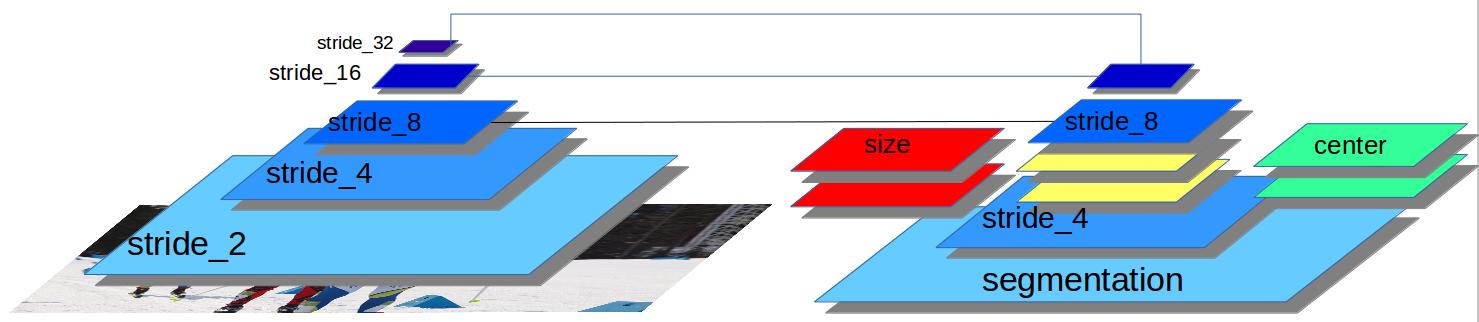}
    \caption{The EOLO framework for instance segmentation.}
\label{structure}
\end{figure*}
 
\subsection{Instance Segmentation}\label{AA}
In FCN\cite{FCN}, the model is trained to predict the score of classification for each pixel. If there is only one object in the picture, the results of semantic segmentation is also instance segmentation. However, if there are two near objects with the same class, FCN cannot distinguish these two object instances. For example, the same pixel can be either the foreground of object 1 or the background of neighboring object 2. In Instance-Sensitive Fully Convolutional Networks (ISFCN)\cite{ISFCN}, the positive-sensitive score map from R-FCN\cite{R-FCN} is proposed to use to instance segmentation. Each score represents the score of a pixel belonging to an object instance at a relative position. The top-level features are used as the input of two full convolution branches, one is used to estimate some instances and generates instance-sensitive score maps it has $k^2$ channels, where $k^2$ is the number of positions; the other branch scores the object for each sliding window.

A sliding window is an ineffective way to extract candidate proposals, FCIS\cite{FCIS} is the first end-to-end instance segmentation network. It continues to use Instance-sensitive score maps, adds inside/outside score maps that distinguish between inside and outside the object instance which introduces the context information. Moreover, the function map and score map in the FCIS network are shared by subsequent object segmentation and detection tasks, however, the previous CNN instance segmentation methods basically run the two tasks of segmentation and detection separately. FCIS is based on box regional proposals like Faster R-CNN, not sliding windows. This method also reduces network parameters and avoids network design choices.

Mask R-CNN continues this kind of top-down method which explores the relationship between pixels and objects in a regional proposal. Use FPN for object detection and semantic segmentation by adding additional branches (extra segmentation branches and original detection branches do not share parameters), thus, MaskR-CNN has three output branches (classification, coordinate regression, and segmentation), the segmentation branch relies on the other two branches results. Mask R-CNN improved the accuracy of RoIpooling. The alignment of candidate regions and convolutional features does not lose information due to quantization through bilinear interpolation. During segmentation, MaskR-CNN decouples the two tasks of determining the category and the output template (mask) and uses sigmoid with the logistic loss function to process each template individually, compared to the classic segmentation Use softmax to make all categories compete better.

ISFCN, FCIS and Mask R-CNN methods segment instance in a bounding box, they fall into the typical top-down paradigm. While bottom-up approaches generate instance masks by grouping the pixels into a set of candidate masks in an image and embed, cluster and assemble them.

Recent YOLACT\cite{yolact} mainly referred to the single-stage detection model RetinaNet. It divided the instance segmentation task into two parallel subtasks: firstly, it generates some prototype masks for each picture through a Protonet network. Secondly, for each instance and bounding box, predict k linear coefficients (Mask Coefficients). Finally, it used a linear combination to generate instance masks. In this process, the network learned how to locate masks with different positions, colors, and semantic instances. SSAP\cite{ssap} proposed a pixel-pair affinity pyramid and according to the affinity of two pixels belonging to the same instance, and sequentially generates instances from coarse to fine by a cascaded graph partition.
Polar Mask and SOLO network both do not belong to bottom-up and top-down paradigms. Polar Mask transfer the pixel classification question to the dense regression question. SOLO projects positions to channels as a dense classification question.

\section{Our Method}
In this section, we first introduce the overall architecture of the EOLO. Then, we redefine the instance segmentation with a 4D vector and semantic segmentation results. Finally, we introduce a new Ellipse Gaussian Kernal Loss function of EOLO.

\subsection{Architecture}
EOLO is a network with a simple structure, which is composed of Mobilenetv3 backbone, feature pyramid network, and three task-specific heads (the center of gravity prediction branch, 4D size prediction branch, and segmentation branches) (Fig. \ref{structure}). The Mobilenetv3 accepts 512$\times$512 inputs with three RGB channels and product feature maps from stride 2 to stride 32, while the feature pyramid network only combines the feature maps from stride 32 to stride 8. These down-sampling final feature maps (high resolution with high dimensional information) are followed by three branches, after the relatively heavy regression processing, we extend low dimensional information (stride 4 and stride 2 feature maps) to improve the performance of segmentation branch. Working in this way, we moved the heavy head from the stride 2 or stride 2 feature maps to stride 8 feature maps, it will save both parameters and computation.

\subsection{Instance segmentation}
In this section, we reformulate object segmentation in a per-pixel prediction approach which follows the up-bottom paradigm. Polar Mask builds up the relationship between a center point and many contour points. Due to Polar Mask defeat the 36 directions from the center to contour, so the Mask boundary could be specific by one value (a distance from center to contour).  The SOLO network builds up the relationship between pixels to cells in object grids. While semantic segmentation cannot distinguish object instances belong to the same classification. This essay finds a new way to distinguish the overlapping parts of objects in the same class, in other words, based on semantic segmentation and object detection, EOLO implemented Instance Segmentation.

EOLO is an up-bottom method, it detects objects and segments semantic first, then combines the semantic segmentation results and object detection results to get the instance segmentation results. 

First, we begin by briefly reviewing the Objects as Points\cite{CenterNet}. Objects as points consist of one single-scale prediction layer $\hat{S} \in R^{\frac{W}{R}\times\frac{H}{R}\times(C+2)}$, where R is the output stride and C is the classes category, 2 represent the scale H and W for an object which center located in a present cell on the heat map. Centernet can adapt for different scales objects because it predicts an object on a high-resolution feature map (stride 4 feature map). While other networks like Faster R-CNN extracts regions of interesting (RoI)\cite{FasterRCNN} from different levels of feature pyramid according to their scale. Generally, they extract three different scales from stride 32,16 and 8 feature maps, they are separately used to predict large, medium and small objects. Stride 4 is more potential to predict all scales object, thus CenterNet keeps this simple structure-only predicts all scales object on stride 4 feature map. This essay aims at applying Instance Segmentation on embedded devices, so we predict all scale objects on single-scale stride 8 feature maps. It effectively reduced computation. But in most models, regarding the center of bounding boxes as the center of an object is unreasonable, in some scenarios, applications require tracking an object by their center of gravity. If we use the center of bounding boxes representing object centers, the results will be ridiculous. Also, using instance masks results to calculate the object gravity center is not a bad idea, but why not directly use the center of gravity as objects' center?

\begin{figure}[htp]
    \centering
    \includegraphics[width=8.5cm]{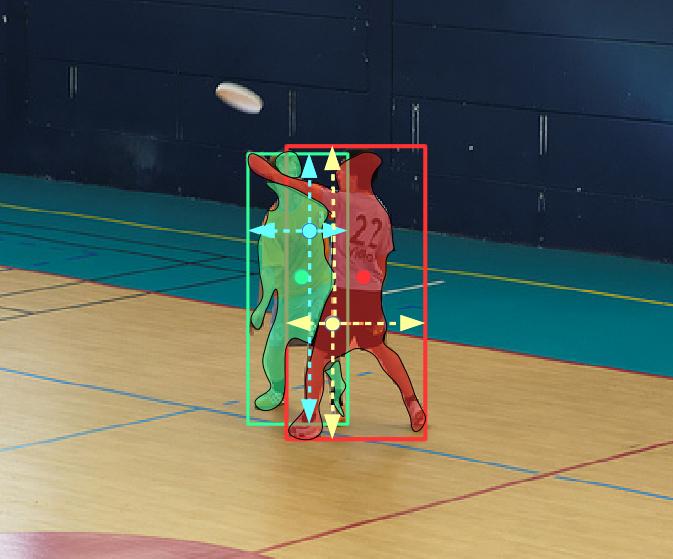}
    \caption{As shown in the image, EOLO works by predicting the center of gravity and a 4D vector (l, t, r, b) encoding the location of a bounding box at each semantic foreground pixel (supervised by object mask information during training). The red rectangular with the red center and the green rectangular with the green center show that there are two objects in the image, but they belong to the same class-person, the semantic segmentation cannot distinguish them, it can be ambiguous in terms of which object this pixel should regress. The key to telling them that is to divide the overlapping parts, EOLO predicts a 4D vector (l, t, r, b) for each pixel on an object. Like the blue and yellow points in the overlapping part, the 4D vectors of these two points show different bounding boxes. Thus, it is easy to classify pixels in overlapping part by calculating the IoU for candidate object bounding box.}
\label{sizemap}
\end{figure}

We want to describe an object with the center of gravity and bounding box, centers with 2D objects size (H, W) is not enough, we need a 4D vector (Left, Top, Right, Bottom) to precisely represent the relationship between center and bounding box. FCOS directly regresses the bounding box of each position in the feature map. FCOS defines Top, Left, Bottom, Right values for each point on the feature maps. In other words, FCOS directly uses each position as a training sample, which is the same as FCN which used for semantic segmentation. But FCOS introduce 4 values to show bounding box, it is do not defining the center of gravity, it is for making sure the bounding box location by each point, and optimized bounding box by Center-Ness layer\cite{FCOS}, which assigns different weights for each point according to whether they are close to the object center. Center-Ness reduces the contribution of inaccurate edge points and increases the contribution of center region points. But in this essay, we regress this 4D vector to directly work on the gravity of the center. This 4D vector also works on each position as a training sample, especially we training EOLO to distinguish the overlapping parts for the same category objects. The size branch not only results in the bounding box size for the center position but also results in each point on an object to label them specifically (Fig. \ref{sizemap}). 

\subsection{Loss function}
\begin{figure}[htp]
    \centering
    \includegraphics[scale=0.2]{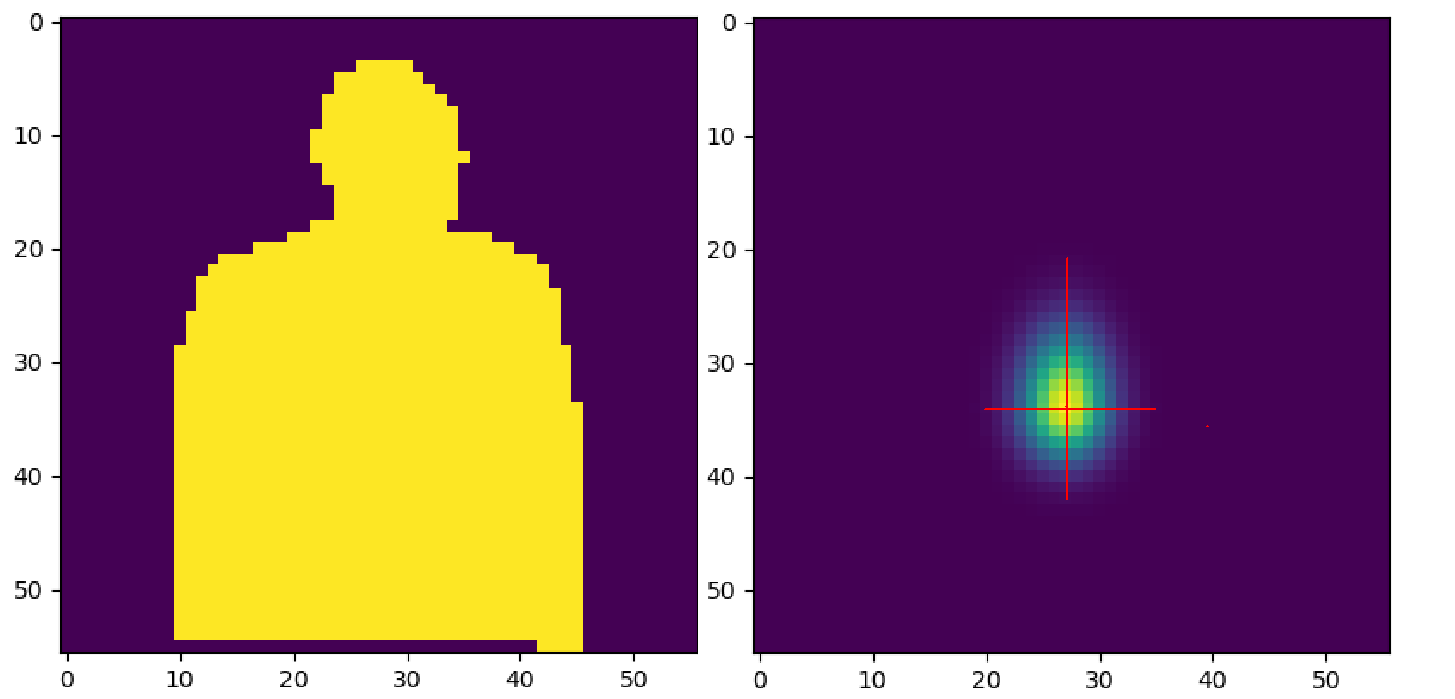}
    \caption{ Ellipse Gaussian Distribution }
\label{masks}
\end{figure}

Finally, we define a multi-task loss for our model as L = L$center$ + L$size$ + L$boundary$ + L$seg$. The center classification loss L$center$ is defined in[]. The center of gravity branch prodnuces a heatmap $\hat{Y}_{center}\in [0,1]^{\frac{W}{R}\times\frac{H}{R}\times C}$, where R is the output stride and C is the number of class types. In this essay we set C = 2 including person and car categories or C = 80 for MS COCO Instance Segmentation task. We use an approximate ellipse Gaussian Distribution kernel
$Y_{xyc}=exp(-\frac{(\frac{x-\Tilde{g_{x}}}{\Tilde{r_x}})^2+(\frac{y-\Tilde{g_{y}}}{\Tilde{r_y}})^2}{2{\sigma_c}^2})$, where $\Tilde{g}\in S^2$ of class C represent each center of gravity, $S^2$ are points at feature map. $\Tilde{r_x}$ is left or right distance from center to boundary. $\Tilde{r_y}$ is top or bottom distance from center to boundary,  we set $\sigma$ as 1, we use scale-transform factors $\Tilde{r_x}$ and $\Tilde{r_y}$ to adapt object size (Fig. \ref{masks}). 

The training target of the center of gravity is to reduce logistic regression with the focal loss[]:
\begin{equation}
L_{c}=\frac{-1}{N}\sum_{xyc}
\left\{
    \begin{array}{lr}
         (1-\hat{Y}_{xyc})^{\alpha}log(\hat{Y}_{xyc})  & {if Y_{xyc}=1} \\ 
         (1-Y_{xyc})^{\beta}(\hat{Y}_{xyc})^{\alpha}\\
         log(1-\hat{Y}_{xyc})) & otherwise\\
    \end{array}
\right.
\end{equation}
$\alpha$ and $\beta$ are hyper-parameters of the focal loss\cite{RetinaNet}, and N is the number of the center of gravity in the input image. This formula slows down the punishment of negative samples around positive samples through $\hat{Y}_{xyc}$ distribution factor, We use $\alpha$=2 and $\beta$=4 in all our experiments, following Law and Deng[]. We give up the local offset loss in[], because offset loss of our experiment has very limited improvement in the accuracy of the center points and it brings much computation.

The size brance produces a heatmap $\hat{Y}_{size}\in [0,\infty]^{\frac{W}{R}\times\frac{H}{R}\times 4}$, where 4 represent 4 distance (right=$d_{r}$, top=$d_{t}$, left=$d_{l}$ and bottom=$d_{b}$ distances) from center of gravity to bounding box. Assume (${x_{1}}^{(k)}$, ${y_{1}}^{(k)}$, ${x_{2}}^{(k)}$, ${y_{1}}^{(k)}$) as the bounding box of object k with category $c_k$. The $p_k$ = (${c_{x}}^(k)$, ${c_{y}}^(k)$) are regarded as the center of gravity of object k with category $c_k$. The distances are ${d_l}^{(k)}$ = ${x_{1}}^{(k)}$ - ${c_{x}}^{(k)}$, ${d_t}^{(k)}$ = ${y_{1}}^{(k)}$ - ${c_{y}}^{(k)}$, ${d_r}^{(k)}$ = ${c_{x}}^{(k)}$ - ${x_{2}}^{(k)}$ and ${d_b}^(k)$ = ${c_{y}}^{(k)}$ - ${y_{2}}^{(k)}$. Basically, we regress to the objects size at each center $s_k$ = ($d_l$, $d_t$, $d_r$, $d_b$). In addition, we regress to point size $s'_k$ = ($d'_l$, $d'_t$, $d'_r$, $d'_b$) for each point as center on feature map. 
$L_{size}$ and $L_{boundary}$ are both working on size branch. We use an L2 loss at center of gravity point:
\begin{equation}
L_{size}=\frac{1}{N}\sum_{k=1}^{N} \ {(\hat{S}_{pk} - S_k)}^2
\end{equation}
\begin{equation}
L_{boundary}=\frac{1}{N}\sum_{k=1}^{N} \ {(\hat{S'}_{pk} - S'_k)}^2
\end{equation}

Our definition of $\hat{Y}_{seg}\in [0,1]^{\frac{W}{R}\times\frac{H}{R}\times C}$ allows the network to generate masks for each class. This network relies on the size branch to predict the object mask, the processing to product class mask is similar with FCNs, but the fashion to decouple the object mask is different from practice like \cite{MaskR-CNN},\cite{FCIS}, \cite{ISFCN},\cite{PolarMask},\cite{SOLO}. In that case, object masks are defined as a position relative question in the local area. In our case, we directly divide contour of objects via $s'_k$, if the positive points on segment branch with the same class belong to the same object, their $s'_k$ should be labeled as the same bounding box, the IoU between center point bounding box and the positive points should achieve high. If they are contributing to different objects, their IoU of the bounding box should be low. It is same with Mask R-CNN avoid cross classes competition, in this essay we use sigmoid activation function and a binary loss:
\begin{equation}
L_{seg}=\frac{-1}{N}\sum_{k=1}^{N}
\left\{
    \begin{array}{lr}
         (1-\hat{S}_{xyc})^{\alpha}log(\hat{S}_{xyc})  & {if S_{xyc}=1} \\ 
         (\hat{S}_{xyc})^{\alpha}log(1-\hat{S}_{xyc}) & otherwise
    \end{array}
\right.
\end{equation}

Here the segmentation branch produces a heatmap $\hat{S}_{xyc}\in [0,1]^{\frac{W}{R}\times\frac{H}{R}\times C}$, where R is the output stride and C is the number of class types. Our framework can be easily extended to panoramic segmentation tasks. We have already modeled the instance segmentation masks for C classes type (for example, person, car), and increase
the classes type as C + K. The additional K categories would only work for semantic segmentation task.
C classes for instance segmentation and K classes for semantic segmentation, extending to Panoramic segmentation will only increase fewer parameters and computation (if C=2 and K=10, increasing less then 2 percent parameters) of the total.

\subsection{Training}
We train on the resolution of input of 512$\times$512. EOLO products an output resolution of 64$\times$64 as intermediate results, then for center and size branches, EOLO uses bi-linear interpolation down-sampling to recover output resolution to 256$\times$256. For the segmentation branch, we use feature maps resolution of 128$\times$128 and 256$\times$256 to detail the results. The intermediate segmentation results heat map fits similar normal distribution, the central area has a high response and the boundary area has a low response, to increase the resolution and accuracy we filter the boundary area by threshold (from 0.5 to 0.8), then blend stride 4 and stride 2 maps to make the boundary precise. We do not use any data augmentation processing to train the model. For the residual MobileNetv3 and YOLO feature pyramid on 80 classes on COCO2017, we train with a batch-size of 8 (on 2 GPU) and learning rate 1e-4 for 120 epochs, with learning rate $\times$10 at 100 epoch. For 2 classes on COCO2017 with a lighter feature pyramid layer and head maps, we train with a batch-size of 16 (on 2 GPU) and learning rate 5e-4 for 120 epochs, with 10$\times$ learning rate dropped at 100 epoch. Heavy version train in 20 days on two RTX 2080ti GPU, while light version requires 4 days.

\begin{table*}
\centering
\caption{Comparison of different Instance Segmentation Methods performance}
\begin{tabular}{ccccccccc}
\hline
 & Backbone & FPS & AP & $AP_{50}$ & $AP_{75}$ & $AP_{S}$ & $AP_{M}$ & $AP_{L}$ \cr
\hline
FCIS & Res-101-C5 & 3 & 29.1 & 49.5 & - & 6.9 & 30.8 & 48.7 \cr
Mask R-CNN & Res-101-FPN & 5 & 38.3 & 58.7 & 38.6 & 16.7 & 16.7 38.7 & 52.1\cr
\hline  %添加表格中横线
TensorMask & Res-101-FPN & - & 34.2 & 56.8 & 36.1 & 15.9 & 36.7 & 48.8\cr
YOLACT & Res-101-FPN & 21 & 30.3 & 50.1 & 31.7 & 11.9 & 32.5 & 42.6\cr 
PolarMask & Res-101-FPN & 3 & 29.1 & 50.4 & 30.6 & 13.3 & 32.4 & 41.7\cr
SOLO & Res-101-FPN & 3 & 37.7 & 59.5 & 40.2 & 15.8 & 40.5 & 53.9 \cr
\hline  %添加表格头部粗线
PolarMask & MobileNetv3-FPN & 8 & 23.8 & 27.6 & 25.1 & 10.6 & 18.4 & 23.7 \cr
SOLO & MobileNetv3-FPN & 8 & 28.8 & 32.3 & 31.7 & 13.4 & 20.2 & 28.9 \cr 
EOLO & MobileNetv3-Conv & 30 & 11.7 & 27.7 & 12.2 & 2.3 & 15.3 & 17.8 \cr
EOLO & MobileNetv3-DPConv & 48 & 6.7 & 21.7 & 9.2 & 1.3 & 12.7 & 7.8\cr
\hline
\end{tabular}
\label{table1}
\end{table*}

\begin{figure*}[h]
\centering
 
\subfigure[]{
    \begin{minipage}[t]{0.2\linewidth}
        \centering
        \includegraphics[width=3.5cm]{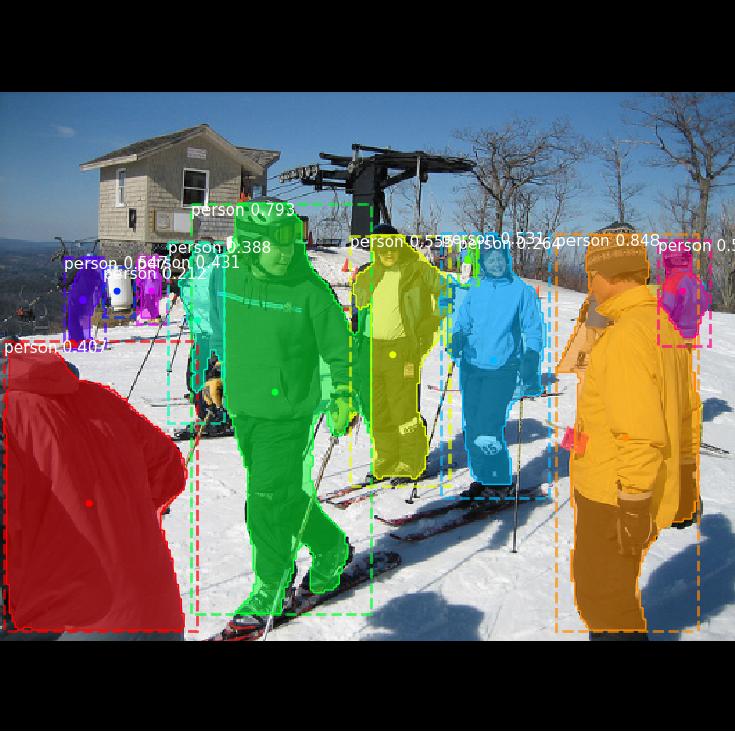}
        %\caption{fig1}
    \end{minipage}%
}%
\subfigure[]{
    \begin{minipage}[t]{0.2\linewidth}
        \centering
        \includegraphics[width=3.5cm]{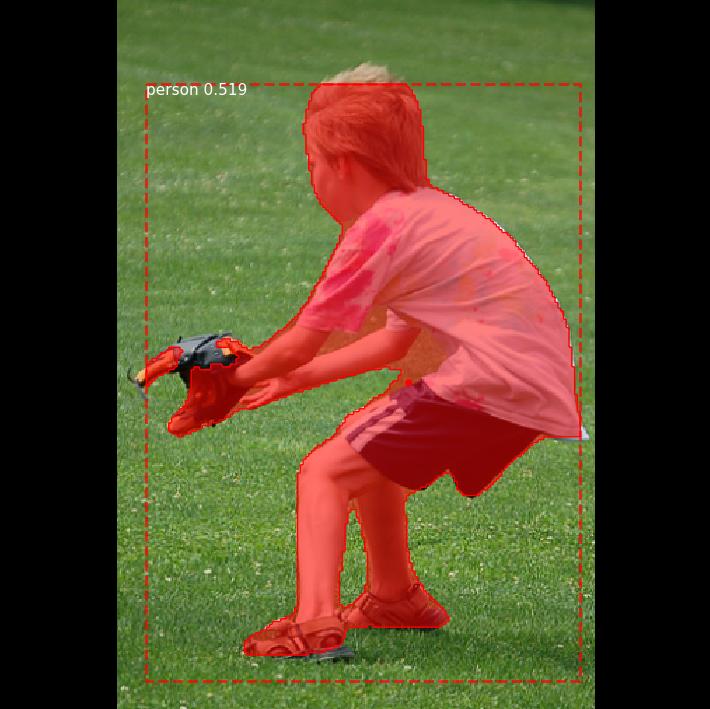}
        %\caption{fig1}
    \end{minipage}%
}%
\subfigure[]{
    \begin{minipage}[t]{0.2\linewidth}
        \centering
        \includegraphics[width=3.5cm]{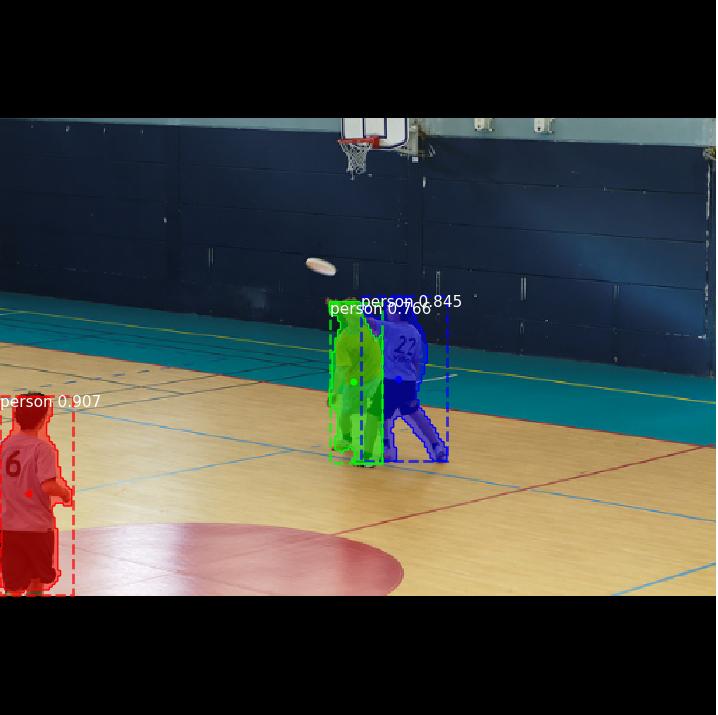}
        %\caption{fig1}
    \end{minipage}%
}%

\subfigure[]{
    \begin{minipage}[t]{0.2\linewidth}
        \centering
        \includegraphics[width=3.5cm]{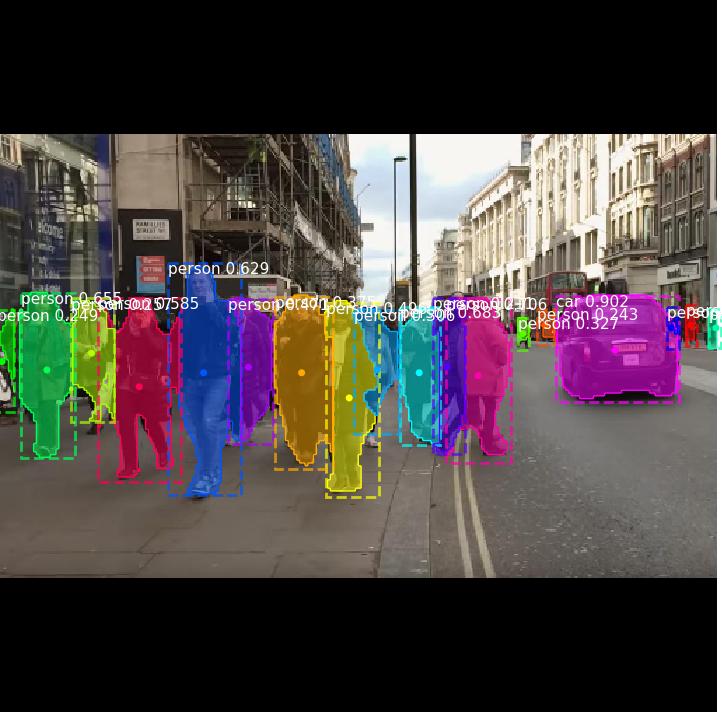}
        %\caption{fig1}
    \end{minipage}%
}%
\subfigure[]{
    \begin{minipage}[t]{0.2\linewidth}
        \centering
        \includegraphics[width=3.5cm]{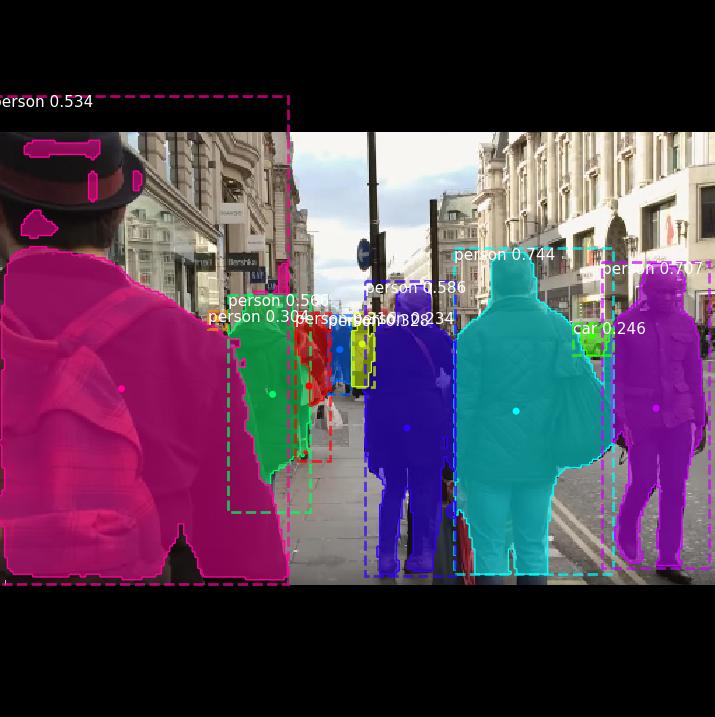}
        %\caption{fig1}
    \end{minipage}%
}%
\subfigure[]{
    \begin{minipage}[t]{0.2\linewidth}
        \centering
        \includegraphics[width=3.5cm]{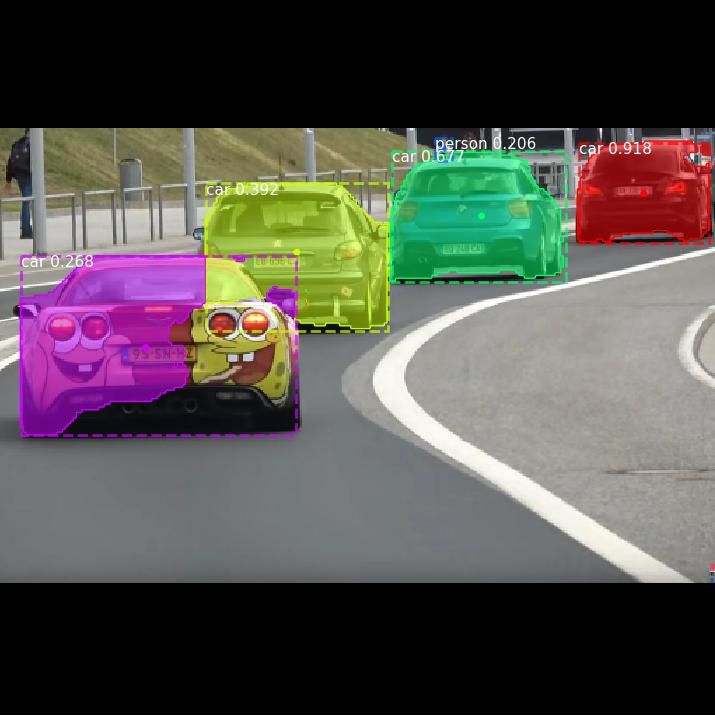}
        %\caption{fig1}
    \end{minipage}%
}%
\centering
\caption{Instance Segmentation results on Rasspberry Pi4.}
\vspace{-0.2cm}
\label{fig:compare_fig}
\end{figure*}

\subsection{Inference}
Anchor-based methods proposed many anchors to improve the recall and they use NMS to reduce the false positive (FP) results.  Because EOLO is an anchor-free method, the inference processing tries to pick the peak value up from the center head map, EOLO does not necessary to use NMS to reduce wrong output. But sometimes EOLO exists repeatable detection in the same area, so NMS will relieve this problem. We do not use deformable convolution kernels\cite{CenterNet}, Ojbect as Points used deformable convolution kernels and it avoids NMS processing. EOLO also needs to compare the IoU in the overlapping area after picking the centers and related bounding boxes up.

\section{Experiments}
We present experimental results on the MS COCO 2017 instance segmentation track. For our main results, we report COCO masks AP on the test-part. We compare EOLO to state-of-the-art methods in instance segmentation on MS COCO test-dev in Table \ref{table1}. EOLO with MobileNetv3 achieves a mask AP of 11.7 and achieves 30 FPS on one 1080Ti GPU. It is not as good as these state-of-the-art methods, but EOLO can conference faster than them. With the depth-and-point wise Convolution replacing normal Convolution Kernel, ELOLO trained on two categories-person and car, this model achieves 48 FPS on 1080Ti. After int8 quantifying it can reach 16 FPS on Raspberry Pi4 with Google Coral USB Accelerator. We tried to use MobileNetv3 as a backbone to rebuild PolarMask and SOLO, from Table \ref{table1} we can see, MobileNetv3 limited the performance of AP, and using Single-Scale head map can not work as well as Multi-Scale head map, especially on Small object, Multi-Scale methods generally have better performance.

\subsection{Ablation Experiments}
As mentioned before, the center of gravity of an object is not the center of the bounding box. The training center with normal circle distribution does not respect object real density distribution. the low-quality center product by circle distribution, it limits the negative sample close around the center of gravity, if the prediction center drifting to the negative sample, it is the low-quality center. Ellipse Gaussian Distribution can according to object size dynamically arrange the punishment range of the close negative sample. As shown in Table \ref{table2} the Ellipse Gaussian Distribution can boost AP from 9.8$\%$ to 11.7$\%$. 

\begin{table}
\centering
\caption{Comparison of different Loss Function Kernel performance}
\begin{tabular}{ccccccc}
\hline  %添加表格头部粗线
$\quad$& AP & $AP_{50}$ & $AP_{75}$ & $AP_{S}$ & $AP_{M}$ & $AP_{L}$ \cr
\hline  %添加表格头部粗线
Circle  & 8.7 & 24.5 & 9.8 & 2.2 & 13.3 & 14.7 \cr
Circle Gaussian  & 9.8 & 25.1 & 10.1 & 2.3 & 13.2 & 14.7\cr
Ellipse Gaussian  & 11.7 & 27.7 & 12.2 & 2.3 & 15.3 & 17.8 \cr
\hline
\end{tabular}
\label{table2}
\end{table}

\section{Conclusion}
In this work we have developed an up-down instance segmentation framework, also refer as EOLO, achieving an acceptable accuracy and great FPS-compare with the same backbone networks. Our proposed model is end-to-end trainable and can instance masks with constant inference time after simple post-processing, eliminating the need for
the grouping post-processing as in bottom-up methods or the RoI operations in top-down approaches.
By introducing the new notion of ‘Ellipse Gaussian’ and '4D Size vector', for
the first time, we can reformulate instance mask prediction into a much-simplified regression task, making instance segmentation significantly simpler than all current approaches. Given the simplicity, flexibility, and acceptable performance of EOLO, we hope that our EOLO can serve as a cornerstone for many instance-level recognition
tasks. Part of the results shows in Fig. \ref{fig:compare_fig}. EOLO performance has greatly improved space, we will continue modifying and improving EOLO.


\end{document}